\documentclass[acmsmall]{acmart}

\AtBeginDocument{%
  \providecommand\BibTeX{{%
    \normalfont B\kern-0.5em{\scshape i\kern-0.25em b}\kern-0.8em\TeX}}}

\setcopyright{acmcopyright}
\copyrightyear{2023}
\acmYear{2023}
\acmDOI{XXXXXXX.XXXXXXX}

\usepackage{makecell}
\usepackage{multirow}
\usepackage{caption}
\usepackage{subcaption}
\usepackage{nicematrix}
\usepackage{pifont}
\usepackage{colortbl}
\usepackage{enumitem}
\usepackage{caption}

\newtheorem{definition}{Definition}
\begin{document}

\title{Class-Imbalanced Learning on Graphs: A Survey}

\author{Yihong Ma}
\email{yma5@nd.edu}
\orcid{0000-0003-4729-5953}
\author{Yijun Tian}
\email{yijun.tian@nd.edu}
\orcid{0000-0001-5410-0566}
\author{Nuno Moniz}
\email{nuno.moniz@nd.edu}
\orcid{0000-0003-4322-1076}
\author{Nitesh V. Chawla}
\email{nchawla@nd.edu}
\orcid{0000-0003-3932-5956}
\affiliation{%
  \institution{University of Notre Dame}
  \city{Notre Dame}
  \state{Indiana}
  \country{USA}
  \postcode{46556}
}

\renewcommand{\shortauthors}{Ma, et al.}

\begin{abstract}
The rapid advancement in data-driven research has increased the demand for effective graph data analysis. However, real-world data often exhibits class imbalance, leading to poor performance of machine learning models. To overcome this challenge, \underline{\textbf{c}}lass-\underline{\textbf{i}}mbalanced \underline{\textbf{l}}earning on \underline{\textbf{g}}raphs (CILG) has emerged as a promising solution that combines the strengths of graph representation learning and class-imbalanced learning. In recent years, significant progress has been made in CILG. Anticipating that such a trend will continue, this survey aims to offer a comprehensive understanding of the current state-of-the-art in CILG and provide insights for future research directions. Concerning the former, we introduce the first taxonomy of existing work and its connection to existing imbalanced learning literature. Concerning the latter, we critically analyze recent work in CILG and discuss urgent lines of inquiry within the topic. 
Moreover, we provide a continuously maintained reading list of papers and code at \href{https://github.com/yihongma/CILG-Papers}{https://github.com/yihongma/CILG-Papers}.
\end{abstract}




\maketitle

\section{Introduction}
\label{sec:introduction}
Graphs are a prevalent and powerful data structure for representing complex relational systems, such as social networks, citation networks, and knowledge graphs. In these systems, nodes symbolize entities, while edges denote their relationships. In recent years, graph representation learning techniques have proven effective in discovering meaningful vector representations of nodes, edges, or entire graphs, resulting in successful applications across a wide range of downstream tasks \cite{tian2022recipe2vec,ma2022hierarchical,yang2016revisiting}. However, graph data often presents a significant challenge in the form of class imbalance, where one class's instances significantly outnumber those of other classes. This imbalance can lead to suboptimal performance when applying machine learning techniques to graph data.

Class-imbalanced learning on graphs (CILG) is an emerging research area addressing class imbalance in graph data, where traditional methods for non-graph data might be unsuitable or ineffective for several reasons. Firstly, graph data's unique, irregular, non-Euclidean structure complicates traditional class-imbalance techniques designed for Euclidean data \cite{zhao2021graphsmote}. Secondly, graph data often holds rich relational information, necessitating specialized techniques for preservation and leverage during the learning process \cite{hgmae}. Lastly, node dependencies and interactions in a graph make class re-balancing complex, as naïve oversampling or undersampling may disrupt the graph's structure and thus lead to poor performance \cite{park2021graphens}.

This survey aims to raise awareness within the graph machine learning community about the expanding field of class-imbalanced learning, which has attracted significant attention in areas such as computer vision and natural language processing. With growing interest and research in this domain, now is an opportune time for a comprehensive survey paper that introduces the background and motivation of CILG, provides an overview of existing CILG techniques under different taxonomies, and presents a thorough evaluation framework for CILG, including benchmark datasets and a critical assessment of current performance metrics.

Moreover, this survey highlights three future research directions of CILG that require further exploration and development to broaden the applicability of class-imbalanced learning techniques in graph-based domains. These directions include investigating CILG techniques beyond node classification tasks, addressing CILG in complex and heterophilous graphs, and exploring CILG in the context of topology imbalance, where the uneven distribution of structural patterns or subgraphs within the graph can lead to biased learning of graph representations.

The paper is organized as follows: Section \ref{sec:prelim} introduces three challenges of CILG and provides background on graph representation learning (GRL) and class-imbalanced learning (CIL). Section \ref{sec:methods} details recent CILG methods, including data-level methods (Section \ref{subsec:data}) and algorithm-level methods (Section \ref{subsec:algo}). Section \ref{sec:eval} provides information on datasets, train-test split strategies, and evaluation metrics used in CILG literature. Finally, Section \ref{sec:future} discusses three future research directions.

To summarize, the main contributions of this paper include:
\begin{itemize}[leftmargin=*]
    \item This paper is the first comprehensive survey on class-imbalanced learning on graphs (CILG);
    \item We systematically categorize CILG methods into two primary categories and further divide them into six sub-categories;
    \item We discuss three pressing future research questions in the CILG domain, providing a roadmap for future exploration in this field;
    \item We maintain a reading list featuring a comprehensive collection of relevant papers and code, which is continuously updated at \href{https://github.com/yihongma/CILG-Papers}{https://github.com/yihongma/CILG-Papers}.
\end{itemize}

\section{Preliminaries}
\label{sec:prelim}
Consider an attributed graph $G=(\mathcal{V},\mathcal{E})$ with a set of $n$ nodes $\mathcal{V} = \{ v_1, \ldots, v_n \}$ and a set of $m$ edges $\mathcal{E} \subseteq \mathcal{V} \times \mathcal{V}$. The node feature matrix $\mathbf{X}\in\mathbb{R}^{n\times d}$ has each row $\mathbf{X}_i$ representing the feature vector of node $v_i$ with a feature dimension of $d$. For node classification tasks, $\mathbf{y} = \{y_1, \ldots, y_n\}$ denotes the set of $n$ node labels, where $\mathbf{y}_i$ corresponds to node $v_i$'s label with $K$ possible class labels. For graph classification tasks, $\mathcal{G} = \{ G_1, \ldots, G_N \}$ represents the set of $N$ graphs, and $\mathbf{y} = \{y_1, \ldots, y_N\}$ denotes the set of $N$ graph labels with $\mathbf{y}_i$ being the label of graph $G_i$ with $K$ potential class labels.

\subsection{Graph Representation Learning}
\label{subsec:grl}
Graph representation learning (GRL) \cite{zhang2020deep} focuses on discovering meaningful vector representations of nodes, edges, or entire graphs for various graph mining applications. Generally, GRL approaches fall into three categories: (1) network embedding models \cite{cui2018survey} that preserve proximities among contextual nodes, capturing the graph structure, (2) graph neural networks (GNNs) \cite{wu2020comprehensive} that aggregate neighboring nodes’ feature information for learning node embeddings, and (3) knowledge graph embedding methods \cite{wang2017knowledge} that learn node and edge embeddings by computing the acceptability score of fact triplets, treating the graph as a collection of such triplets.

GNNs, as the current state-of-the-art of GRL, are the dominant backbone for CILG methods. GNNs typically adopt a message-passing mechanism through neighborhood aggregation, updating a node's representation by aggregating information from its neighboring nodes and edges. After $k$ iterations of aggregation, a node's representation encapsulates the structural information within its $k$-hop neighborhood, which is defined as:
\begin{equation}
\label{eq:gnn}
    \mathbf{h}_v^{(k)} = \textsc{Update} (\mathbf{h}_v^{(k-1)}, \textsc{Aggregate} (\{ \mathbf{h}_u^{(k-1)} \mid \forall u \in \mathcal{N}(v) \})),
\end{equation}
where $\mathbf{h}_v^{(k)}$ is the representation vector of node $v \in \mathcal{V}$ in the $k$-th GNN layer; $\mathcal{N}(v)$ represents the set of neighbors of node $v$; $\textsc{Aggregate}(\cdot)$ is the neighbor aggregation function and $\textsc{Update}(\cdot)$ is the combination function. $\mathbf{h}_v^{(0)}$ is initialized with node attribute $\mathbf{X}_v$. 

\subsection{Class-Imbalanced Learning}
\label{subsec:cil}
Class-imbalanced learning (CIL) \cite{haixiang2017learning,johnson2019survey,lopez2013insight,branco2016survey} aims to address the problem of learning from imbalanced class distributions, where certain classes (i.e., majority classes) have much more training instances than others (i.e., minority classes). 
Given a set of training samples $\mathbf{x} = \{\mathbf{x}_1, \ldots, \mathbf{x}_n\}$, and their corresponding class labels $\mathbf{y} = \{y_1, \ldots, y_n\}$, let $\mathcal{C}_k$ be the set of labeled samples in class $k$ with $k \in \{ 1, \ldots, K \}$. The class-imbalanced problem occurs when the class distribution $P_k$ is highly skewed. To quantify the severity of class imbalance, we use an imbalance ratio $\rho$ -- the ratio between the number of samples in the head and tail classes. Both $P_k$ and $\rho$ are defined as follows:
\begin{equation}
P_k = \frac{|\mathcal{C}_k|}{\sum_{i=1}^{K} |\mathcal{C}_i|}, \quad
\rho = \frac{\max_k |\mathcal{C}_k|}{\min_k |\mathcal{C}_k|}.
\end{equation}

Traditional CIL methods generally fall into two categories: (1) data-level methods that aim to modify the distribution of the training data itself to balance the class distribution, such as over-sampling the minority class \cite{chawla2002smote,he2008adasyn,mullick2019generative}, under-sampling the majority class \cite{laurikkala2001improving,mani2003knn}, and hybrid-sampling \cite{ramentol2012smote,saez2015smote} that combines both; and (2) algorithm-level methods that aim to modify the learning algorithm itself to better handle the class imbalance, such as cost-sensitive learning \cite{tang2008svms,zhou2005training}, ensemble learning \cite{chawla2003smoteboost,guo2004learning,seiffert2009rusboost}, 
and loss function engineering \cite{lin2017focal,cui2019class,cao2019learning,dong2018imbalanced}.

\subsection{Class-Imbalanced Learning on Graphs}
Class-imbalanced learning on graphs (CILG) refers to learning from imbalanced graphs where the distribution of classes is skewed, making it difficult to train models that can effectively classify nodes, predict relations, or classify the entire graph. Only a few studies \cite{zhang2022diving,wang2022imbalanced} have explored the problem of graph classification, while the majority have focused on node classification -- the focus of this survey. We formally define the two graph learning tasks under the class-imbalanced setting:

\begin{definition}[Class-Imbalanced Node Classification]
Given a graph $G$ and a set of labeled nodes $\mathcal{V}^\ell \subseteq \mathcal{V}$ that are class-imbalanced, the goal is to learn a classifier $f$ that assigns a class label to each unlabeled node in $\mathcal{V}^u = \mathcal{V} \setminus \mathcal{V}^\ell$ with high accuracy in both majority and minority classes.
\end{definition}

\begin{definition}[Class-Imbalanced Graph Classification]
Given a set of labeled graphs $\mathcal{G}^\ell \subseteq \mathcal{G}$ that are class-imbalanced, the goal is to learn a classifier $f$ that assigns a class label to each unlabeled graph in $\mathcal{G}^u = \mathcal{G} \setminus \mathcal{G}^\ell$ with high accuracy in both majority and minority classes.
\end{definition}

\subsection{Motivation and Challenges}
CILG has recently emerged as a significant research area due to the prevalence of imbalanced class distributions in graph data. In many real-world applications, class imbalance is a common and severe issue. The imbalance can cause standard machine learning models to favor the majority class, leading to poor performance on the minority class. This is particularly detrimental in graph data, where dependence and correlation among data points can lead to cascading errors and a significant impact on downstream applications. Thus, studying CILG is critical to ensure the reliability and fairness of machine learning models on graph data and enable real-world applications with accurate and unbiased predictions. However, the problem of CILG faces three major challenges:

\begin{itemize}[leftmargin=*]
\item \textbf{The uniqueness of the CILG problem.}
One of the challenges of CILG stems from the unique blend of difficulties in graph representation learning (GRL) and class-imbalance learning (CIL). Additionally, graph data's non-Euclidean characteristics add further difficulty.

\item \textbf{The complexity of methodologies.} 
The efficacy of a CILG approach is heavily dependent on the effectiveness of data preprocessing, the selection of model architecture, and the appropriate design of re-balancing techniques. 

\item \textbf{Lack of a fine-grained taxonomy.}
Currently, CILG methods are classified as data-level or algorithm-level. A finer-grained taxonomy would help better describe existing techniques.
\end{itemize}

\section{Methods}
\label{sec:methods}

\begin{table}[t]
\caption{A comprehensive overview of representative CILG models, arranged in chronological order.}
\label{tab:models}
\begin{center}
\resizebox{\textwidth}{!}
{\begin{NiceTabular}{c|c|c|c|c|c|c|c}

\toprule
\rowcolor[gray]{.9}\multicolumn{8}{c}{\textbf{Data-Level}} \\
\midrule
\textbf{Model} & \textbf{Year} & \textbf{Venue} & \textbf{\makecell{Data\\ Interpolation}} & \textbf{\makecell{Adversarial\\ Generation}} & \textbf{\makecell{Pseudo-\\ Labeling}} & \textbf{Key Components} \\
\midrule
SPARC \cite{zhou2018sparc} & 2018 & KDD & & & \ding{51} & Label propagation; Self-paced learning & \href{http://www.google.com/url?q=http\%3A\%2F\%2Fpublish.illinois.edu\%2Fdaweizhou\%2Ffiles\%2F2019\%2F10\%2FSPARC.zip\&sa=D\&sntz=1\&usg=AOvVaw0Ua6DhuOfhYWplG0XNAlZl}{link} \\
GraphSMOTE \cite{zhao2021graphsmote} & 2021 & WSDM & \ding{51} & & & SMOTE; Pre-training & \href{https://github.com/TianxiangZhao/GraphSmote}{link} \\
GraphENS \cite{park2021graphens} & 2021 & ICLR & \ding{51} & & & mixup; Neighbor sampling; Saliency filtering & \href{https://github.com/JoonHyung-Park/GraphENS}{link} \\
ImGAGN \cite{qu2021imgagn} & 2021 & KDD & & \ding{51} & & Graph structure reconstruction & \href{https://github.com/Leo-Q-316/ImGAGN}{link} \\
D-GCN \cite{wang2021dual} & 2021 & CSAE & & & & $k$-NN; Graph structure generator & - \\
SET-GNN \cite{juan2021exploring} & 2021 & ICONIP & & & \ding{51} & Label propagation; Self-training & - \\ 
GraphMixup \cite{wu2022graphmixup} & 2022 & ECML/PKDD & \ding{51} & & & mixup; Auxiliary objectives & \href{https://github.com/LirongWu/GraphMixup}{link} \\
GATSMOTE \cite{liu2022gatsmote} & 2022 & Mathematics & \ding{51} & & & SMOTE; Attention & - \\
SORAG \cite{duananonymity} & 2022 & ECML/PKDD & & \ding{51} & & Node generator; Edge generator & - \\
DPGNN \cite{wang2022distance} & 2022 & MLG & & & \ding{51} & Label propagation; Metric learning  & \href{https://github.com/YuWVandy/DPGNN}{link} \\
GNN-CL \cite{li2022graph} & 2022 & arXiv & \ding{51} & & & SMOTE; Curriculum learning; Attention & - \\
GraphSR \cite{zhou2023graphsr} & 2023 & AAAI & & & \ding{51} & Label propagation; Reinforcement learning & - \\
\midrule

\rowcolor[gray]{.9}\multicolumn{8}{c}{\textbf{Algorithm-Level}} \\
\midrule
\textbf{Model} & \textbf{Year} & \textbf{Venue} & \textbf{\makecell{Model\\ Refinement}} & \textbf{\makecell{Loss Function\\ Engineering}} & \textbf{\makecell{Post-hoc\\ Adjustments}} & \textbf{Key Components} \\
\midrule
RSDNE \cite{wang2018rsdne} & 2018 & AAAI & \ding{51} & & & Random walk; Intra/Inter-class similarity & \href{https://github.com/zhengwang100/RSDNE-python}{link} \\
ImVerde \cite{wu2018imverde} & 2018 & Big Data & \ding{51} & & & Random walk; Balanced sampling & \href{https://github.com/jwu4sml/ImVerde}{link} \\
DR-GCN \cite{shi2020multi} & 2020 & IJCAI & & & \ding{51} & Adversarial training; Distribution alignment & \href{https://github.com/codeshareabc/DRGCN}{link} \\
RECT \cite{wang2020network} & 2020 & TKDE & \ding{51} & & & Class-label relaxation & \href{https://github.com/zhengwang100/RECT}{link} \\
ReNode \cite{chen2021topology} & 2021 & NeurIPS & & \ding{51} & & Topology imbalance; Node reweighting & \href{https://github.com/victorchen96/ReNode}{link} \\
FRAUDRE \cite{zhang2021fraudre} & 2021 & ICDM & & \ding{51} & & Imbalanced distribution-oriented loss & \href{https://github.com/FraudDetection/FRAUDRE}{link} \\
TAM \cite{song2022tam} & 2022 & ICLR & & \ding{51} & & Topology-aware margin loss & \href{https://github.com/Jaeyun-Song/TAM}{link} \\
CM-GCL \cite{qian2022comodality} & 2022 & NeurIPS & \ding{51} & \ding{51} & & Contrastive learning; Network pruning & \href{https://github.com/graphprojects/CM-GCL}{link} \\
LTE4G \cite{yun2022lte4g} & 2022 & CIKM & \ding{51} & & \ding{51} & Degree imbalance; Knowledge distillation & \href{https://github.com/SukwonYun/LTE4G}{link} \\
ACS-GNN \cite{ma2022attention} & 2022 & ICNSC & \ding{51} & \ding{51} & & Cost-sensitive learning; Attention & - \\
EGCN \cite{wang2022effective} & 2022 & ICNSC & \ding{51} & & & Class-weighted aggregation & - \\
KINC-GCN \cite{bai2022kernel} & 2022 & ICNSC & \ding{51} & & & Kernel propagation; Node clustering & - \\
FACS-GCN \cite{santos2022facs} & 2022 & IJCNN & & \ding{51} & & Cost-sensitive learning; Adversarial training & \href{https://github.com/frsantosp/FACS-GCN}{link} \\
GraphDec \cite{zhang2022diving} & 2022 & GLFrontiers & \ding{51} & \ding{51} & & Contrastive learning; Sparsity training & \href{https://www.dropbox.com/sh/8jaq9zekzl3khni/AAA0kNDs_UMxj4YbTEKKyiXna?dl=0}{link} \\
ImGCL \cite{zeng2022imgcl} & 2023 & AAAI & \ding{51} & \ding{51} & & Contrastive learning; Balanced sampling & - \\
\bottomrule

\end{NiceTabular}}
\end{center}
\end{table}

This section reviews existing CILG approaches, categorizing them into two main groups: (1) data-level methods and (2) algorithm-level methods. Data-level methods are further divided into (i) data interpolation, (ii) adversarial generation, and (iii) pseudo-labeling. Algorithm-level methods are classified into (i) model refinement, (ii) loss function engineering, and (iii) post-hoc adjustments. Table \ref{tab:models} presents an extensive summary of CILG models.

\subsection{Data-Level Methods}
\label{subsec:data}
Data-level techniques are crucial for addressing class imbalance issues in CIL by modifying the training data in either feature or label spaces to achieve a more balanced learning environment. Basic data-level approaches include over-sampling, which increases the number of minority class samples, and under-sampling, which decreases the majority class samples. However, new techniques are required to augment training data within the graph structural space due to node dependencies and interconnections in graph data. In CILG tasks, we further classify data-level methods into data interpolation, adversarial generation, and pseudo-labeling. Each method differs in its approach to generating features, structures, or labels for synthetically created minority data instances.

\subsubsection{Data Interpolation}
\label{subsubsec:int}
Data interpolation generates synthetic training samples for minority classes by forming linear combinations of existing data samples. The Synthetic Minority Over-sampling Technique (SMOTE) \cite{chawla2002smote} is a simple and effective data interpolation method for addressing class imbalance. SMOTE produces a virtual minority training sample $\tilde{\mathbf{x}}$ in the feature space by interpolating two labeled training instances from the same minority class:
\begin{equation}
    \tilde{\mathbf{x}} = \mathbf{x}_i + \lambda (\mathbf{x}_j - \mathbf{x}_i),
\end{equation}
where $\mathbf{x}_i$ is a random minority instance; $\mathbf{x}_j$ is one of the $k$-nearest neighbors of $\mathbf{x}_i$ from the same minority class; $\lambda \in [0, 1]$. 
However, directly applying SMOTE to graphs is challenging as it requires the interpolation process also to consider the topological structure of graph data. To address this issue, GraphSMOTE \cite{zhao2021graphsmote} was proposed as the first data interpolation method on graphs by generating a synthetic minority node through interpolation between two real minority nodes in the embedding space. GraphSMOTE pre-trains an edge predictor using a graph reconstruction objective on real nodes and existing edges to determine the connectivity between the synthetic node and existing nodes.
GATSMOTE \cite{liu2022gatsmote} and GNN-CL \cite{li2022graph} further employ attention mechanisms in their edge generators to enhance predicted edge quality between synthetic and real nodes.

Another common data interpolation technique, mixup \cite{zhang2017mixup}, performs linear interpolation on data instances across all classes instead of focusing solely on minority classes. In particular, mixup creates a synthetic training sample $(\tilde{\mathbf{x}}, \tilde{\mathbf{y}})$ in both feature and label spaces:
\begin{equation}
    \tilde{\mathbf{x}} = \lambda \mathbf{x}_i + (1 - \lambda) \mathbf{x}_j, \quad 
    \tilde{\mathbf{y}} = \lambda \mathbf{y}_i + (1 - \lambda) \mathbf{y}_j,
\end{equation}
where $(\mathbf{x}_i, \mathbf{y}_i)$ and $(\mathbf{x}_j, \mathbf{y}_j)$ are two random instances, and $\lambda \in [0, 1]$.
To address the neighbor memorization issue in GraphSMOTE and its variants, GraphENS \cite{park2021graphens} generates a synthetic minority node using mixup between a real minority node and a random target node. GraphENS filters out class-specific attributes of the target node using a gradient-based feature saliency that measures the importance of each feature in predicting the target node's label.
GraphMixup \cite{wu2022graphmixup}, on the other hand, performs mixup in the semantic space rather than the input or embedding space, preventing the generation of out-of-domain minority samples. GraphMixup incorporates two auxiliary self-supervised learning objectives: local-path prediction and global-path prediction.


\subsubsection{Adversarial Generation} 
\label{subsubsec:adv}
Adversarial generation methods utilize generative adversarial networks (GANs) variants to create synthetic minority nodes. GANs \cite{goodfellow2020generative} are a class of neural networks comprising a generator and a discriminator. The generator \emph{G} aims to create fake data that closely resembles real data, while the goal of discriminator \emph{D} is to differentiate between the real and the generated data. GANs are trained using a minimax objective function, defined as:
\begin{equation}
    \min_{G}\max_{D}\mathbb{E}_{x\sim p_{\text{data}}(x)}[\log{D(x)}] +  \mathbb{E}_{z\sim p_{\text{z}}(z)}[\log{(1 - D(G(z)))}],
\end{equation}
where $p_{data}$ is the distribution of real data $x$, and $p_z$ is the distribution of random noise $z$. In CILG, the generator synthesizes minority nodes, and the discriminator ensures that generated nodes resemble real minority nodes.
ImGAGN \cite{qu2021imgagn}, the first to apply adversarial generation to graphs, introduces a generator network for synthesizing minority nodes and their links to real minority nodes. It derives features from averaging neighboring minority node features. A GCN is used as the discriminator to differentiate between real and synthetic nodes and classify whether they belong to the minority class. ImGAGN is limited to binary classification, and extending it to multi-class classification is non-trivial -- it requires a separate generator for each class.
SORAG \cite{duananonymity}, on the other hand, targets multi-class node classification in graphs by proposing an ensemble of a GAN and a conditional GAN (cGAN) as the node generator. The GAN generates unlabeled synthetic minority nodes and the cGAN creates labeled synthetic minority nodes. The connectivity between synthetic and real nodes is based on the weighted inner product of their node features.


\subsubsection{Pseudo-Labeling}
\label{subsubsec:pseudo}
In graph-based tasks, training data is often partially labeled due to the high cost of human annotation, making semi-supervised learning settings more practical. Pseudo-labeling leverages the large number of unlabeled nodes in graphs to utilize more information without introducing new data instances, distinguishing it from data interpolation or adversarial generation methods. Traditional approaches for generating pseudo labels in graphs include label propagation \cite{zhu2002learning,zhu2005semi}, which iteratively propagates a node's label to neighboring nodes based on their proximity. More recent approaches, such as SPARC \cite{zhou2018sparc} and SET-GNN \cite{juan2021exploring}, incorporate pseudo-labeling into self-training to enrich minority class training samples. In contrast, DPGNN \cite{wang2022distance} transfers knowledge from head classes to tail classes using learned class prototypes and metric learning. Meanwhile, GraphSR \cite{zhou2023graphsr} generates pseudo-labels for unlabeled nodes using a GNN and identifies the most reliable and informative unlabeled nodes based on their similarity to labeled nodes, adaptively enriching minority class training nodes through reinforcement learning.


\subsection{Algorithm-Level Methods}
\label{subsec:algo}
Algorithm-level methods aim to modify learning algorithms to tackle class imbalance. These approaches adjust the model's training process to mitigate problems arising from imbalanced class distribution. We classify CILG algorithm-level methods into three categories: model refinement, loss function engineering, and post-hoc adjustments, depending on the alteration to learning algorithms.

\subsubsection{Model Refinement}
\label{subsubsec:model}
Model refinement involves adapting the underlying architecture of graph representation learning methods to improve their performance in dealing with class imbalance. This can be accomplished through various approaches, such as altering the network's overall structure, incorporating specific modules designed to handle class imbalance, or modifying the learning paradigm. Model refinement has been applied to both network embedding methods and graph neural networks (GNNs) in the context of CILG.

For network embedding methods, RSDNE \cite{wang2018rsdne} and ImVerde \cite{wu2018imverde} modify DeepWalk \cite{perozzi2014deepwalk}, a method that preserves neighborhood structure based on random walks. RSDNE adds auxiliary learning objectives to ensure that the embedding space reflects both intra-class similarity and inter-class dissimilarity, while ImVerde adjusts the transition probability during random walks to encourage minority nodes to stay within the same class, creating better class-separable representations. ImVerde also employs context and balanced-batch sampling to sample node-context pairs based on label information and network topology in a class-balanced manner.

In contrast, ACS-GNN \cite{ma2022attention} and EGCN \cite{wang2022effective} modify the aggregation operation $\textsc{Aggregate}(\cdot)$ in a standard GNN architecture shown in Eq. (\ref{eq:gnn}).
ACS-GNN uses an attention mechanism to assign personalized weights to minority and majority samples, while EGCN limits the aggregation of inter-class edges from a local perspective using estimation density and focuses more on the minority class based on the imbalance ratio from a global perspective.
Besides modifying the standard components of a GNN model, KINC-GCN \cite{bai2022kernel} introduces two modules to enhance node embeddings and exploit higher-order structural features as a preprocessing step before applying GNN for classification. The first is a self-optimizing cluster analysis module that performs clustering on the obtained node embeddings. The second module, the graph reconstruction module, uses an inner product decoder to reconstruct original graphs through reconstructed embedding vectors.

\subsubsection{Loss Function Engineering}
\label{subsubsec:loss}
Loss function engineering is a technique to design tailored loss functions to address class imbalance in machine learning models, improving performance in minority classes. This is achieved through two strategies: (1) assigning greater weight to the loss of minority class data samples during training, making the model more sensitive to errors in these classes and improving overall performance \cite{lin2017focal,cui2019class}, or (2) expanding the decision boundary between minority and majority classes, by modifying the loss function to encourage the model to create a wider separation between classes, reducing misclassification and improving generalization \cite{cao2019learning,dong2018imbalanced}. Loss function engineering can be used independently or combined with other class-imbalanced methods to enhance model performance. However, directly applying these methods to node classification under class-imbalanced settings is challenging due to the connectivity properties between nodes in graphs. ReNode \cite{chen2021topology} and TAM \cite{song2022tam} are two recent works in CILG that incorporate graph topology information into their loss function designs.

ReNode, corresponding to the first strategy mentioned earlier, re-weights the influence of labeled nodes based on their relative positions to class boundaries:
\begin{equation}
    \mathcal{L}_{\text{ReNode}} = \frac{1}{|\mathcal{V}^l|} \sum_{v \in \mathcal{V}^l} w_v \frac{|\overline{\mathcal{C}}|}{|\mathcal{C}_{y_v}|} \mathcal{L}(\boldsymbol{l}_v, y_v),
\end{equation}
where $\mathcal{L}(\cdot, \cdot)$ represents the original loss function, and $\boldsymbol{l}_v \in \mathbb{R}^K$ denotes the logit of node $v$. The term $w_v$ is the modified training weight of node $v$, emphasizing nodes closer to topological class centers. $|\overline{\mathcal{C}}|$ is the average number of training samples across all classes, while ${|\mathcal{C}_{y_v}|}$ refers to the number of training samples in the class to which node $v$ belongs.

TAM, belonging to the second strategy, considers the local topology of individual nodes and adaptively adjusts margins for topologically improbable nodes:
\begin{equation}
    \mathcal{L}_{\text{TAM}} = \frac{1}{|\mathcal{V}^l|} \sum_{v \in \mathcal{V}^l} \mathcal{L}(\boldsymbol{l}_v + \alpha m_v^{\text{ACM}}+ \beta m_v^{\text{ADM}}, y_v),
\end{equation}
where $\mathcal{L}(\cdot, \cdot)$ denotes the original loss function, and $\boldsymbol{l}_v \in \mathbb{R}^K$ represents the logit of node $v$. The term $m_v^{\text{ACM}} \in \mathbb{R}^K$ adjusts the margin of each class by calibrating the deviation of the neighbor label distribution for node $v$. The term $m_v^{\text{ADM}} \in \mathbb{R}^K$ modifies the target class margin based on the relative proximity to the target class compared to the self class. $\alpha$ and $\beta$ are two hyperparameters.

Another line of research has emerged from recent efforts to incorporate self-supervised learning into supervised CIL \cite{jiang2021self,yang2020rethinking,jiang2021improving}. By pre-training a feature extractor using self-supervised learning, class imbalance issues can be mitigated due to its regularization effect on creating a more balanced feature space. The most popular self-supervised graph learning method in this context is contrastive learning (CL) \cite{jaiswal2020survey}. This approach leverages structural information and node features to learn informative node representations while preserving the similarity between nodes that are close in the graph structure. Specifically, given a positive pair of nodes $(u, v^+)$ and a set of negative nodes $\mathcal{V}^-$, the most popular InfoNCE loss \cite{oord2018representation} can be defined as:
\begin{equation}
\mathcal{L}_{\text{CL}} = -\sum_{v^- \in \mathcal{V}^-} \log \frac{\exp(\text{sim}(\boldsymbol{e}_u, \boldsymbol{e}_{v^+}) / \tau)}{\exp(\text{sim}(\boldsymbol{e}_u, \boldsymbol{e}_{v^+}) / \tau) + \sum_{v^- \in \mathcal{V}^-} \exp(\text{sim}(\boldsymbol{e}_u, \boldsymbol{e}_{v^-}) / \tau)},
\end{equation}
where $\boldsymbol{e}_v$ represents the embedding of node $v$; $\text{sim}(\cdot, \cdot)$ is a similarity function that computes the similarity between two embeddings; $\tau$ is a temperature parameter that controls the concentration of the probability distribution.
Recent works exploring the benefits of self-supervised learning on CILG include CM-GCL \cite{qian2022comodality}, GraphDec \cite{zhang2022diving}, and ImGCL \cite{zeng2022imgcl}.

\subsubsection{Post-hoc Adjustments}
\label{subsubsec:posthoc}
Post-hoc adjustments refer to a set of techniques applied after the primary training process of a machine learning model to enhance its performance on imbalanced datasets. These methods focus on fine-tuning, recalibrating, or aligning model outputs during the inference phase or late stages of training rather than altering the data or training procedure itself. By addressing issues such as class imbalance, confidence calibration, and distribution shifts, post-hoc adjustments contribute to improved prediction accuracy, reliability, and generalization in real-world scenarios where data may exhibit significant disparities between classes or other structural imbalances. 
This category of methods remains relatively under-explored within the context of CILG. However, two prominent models in this category are DR-GCN \cite{shi2020multi} and LTE4G \cite{yun2022lte4g}. DR-GCN incorporates a distribution alignment module that ensures unlabeled nodes follow a similar latent distribution to labeled nodes by minimizing the distribution difference based on Kullback-Leibler divergence. In contrast, LTE4G employs a class prototype-based inference method to adjust predictions after the model's training is complete during the inference phase. This approach involves computing the prototype vector for each class by averaging the embeddings of all nodes in that class. The test node's class is then predicted by calculating its similarity with each class prototype vector and assigning the class with the highest similarity score.

\section{Evaluation}
\label{sec:eval}
\begin{table}[t]
\caption{A comprehensive overview of benchmark datasets for real-world node classification on graphs, including information on their application domains, imbalance ratios $\rho$, and label distributions (\%). Datasets are organized by their application domains and presented in chronological order.}
\label{tab:datasets}
\begin{center}
\resizebox{\textwidth}{!}
{\begin{NiceTabular}{c|c|c|cccccccccc}
\toprule
\textbf{Domain} & \textbf{Dataset} & $\boldsymbol{\rho}$ & \textbf{L$_0$} & \textbf{L$_1$} & \textbf{L$_2$} & \textbf{L$_3$} & \textbf{L$_4$} & \textbf{L$_5$} & \textbf{L$_6$} & \textbf{L$_7$} & \textbf{L$_8$} & \textbf{L$_9$}\\ 
\midrule
\multirow{3}{*}{Citation network}
& Cora \cite{yang2016revisiting} & 5 & 30.21 & 15.73 & 15.43 & 12.96 & 11.00 & 8.01 & 6.65 & - & - & - \\
& Citeseer \cite{yang2016revisiting} & 3 & 21.07 & 20.08 & 17.91 & 17.73 & 15.26 & 7.94 & - & - & - & - \\
& PubMed \cite{yang2016revisiting} & 2 & 39.94 & 39.25 & 20.81 & - & - & - & - & - & - & - \\
\midrule
\multirow{2}{*}{Co-purchase network}
& Amazon-Photo \cite{shchur2018pitfalls} & 6 & 25.37 & 22.04 & 11.96 & 11.53 & 10.76 & 9.19 & 4.82 & 4.43 & - & -\\
& Amazon-Computers \cite{shchur2018pitfalls} & 18 & 37.51 & 15.68 & 15.58 & 10.28 & 5.95 & 3.94 & 3.54 & 3.17 & 2.24 & 2.12 \\
\midrule
\multirow{3}{*}{Social network}
& Flickr \cite{Zeng2020GraphSAINT:} & 10 & 40.26 & 25.73 & 9.53 & 7.19 & 5.90 & 5.49 & 3.90 & - & - & - \\
& GitHub \cite{rozemberczki2021multi} & 3 & 74.17 & 25.83 & - & - & - & - & - & - & - & - \\
& Facebook \cite{rozemberczki2021multi} & 2 & 30.62 & 28.91 & 25.67 & 14.81 & - & - & - & - & - & - \\
\midrule
Knowledge graph
& Wiki-CS \cite{mernyei2020wiki} & 9 & 22.90 & 18.40 & 16.52 & 12.17 & 7.39 & 6.67 & 5.70 & 4.20 & 3.53 & 2.52 \\
\bottomrule
\end{NiceTabular}}
\end{center}
\end{table}

The following sections provide a critical overview of the evaluation settings for CILG, including benchmark datasets for node classification, class-imbalanced train-test split strategies, and commonly used performance metrics in CILG research.

\subsection{Datasets}
Table \ref{tab:datasets} summarizes real-world benchmark datasets for graph node classification on graphs. The objectives of these classification tasks differ based on their application domains, such as classifying research areas of paper nodes in citation networks and predicting product categories in co-purchase networks. The datasets typically exhibit a class-imbalance property with imbalance ratios ranging from 2 to 18, with the Amazon-Computers dataset \cite{shchur2018pitfalls} having the most skewed label distribution.

\subsection{Class-Imbalanced Train-Test Split}

\begin{figure}[t]
    \centering
    \includegraphics[width=0.9\textwidth]{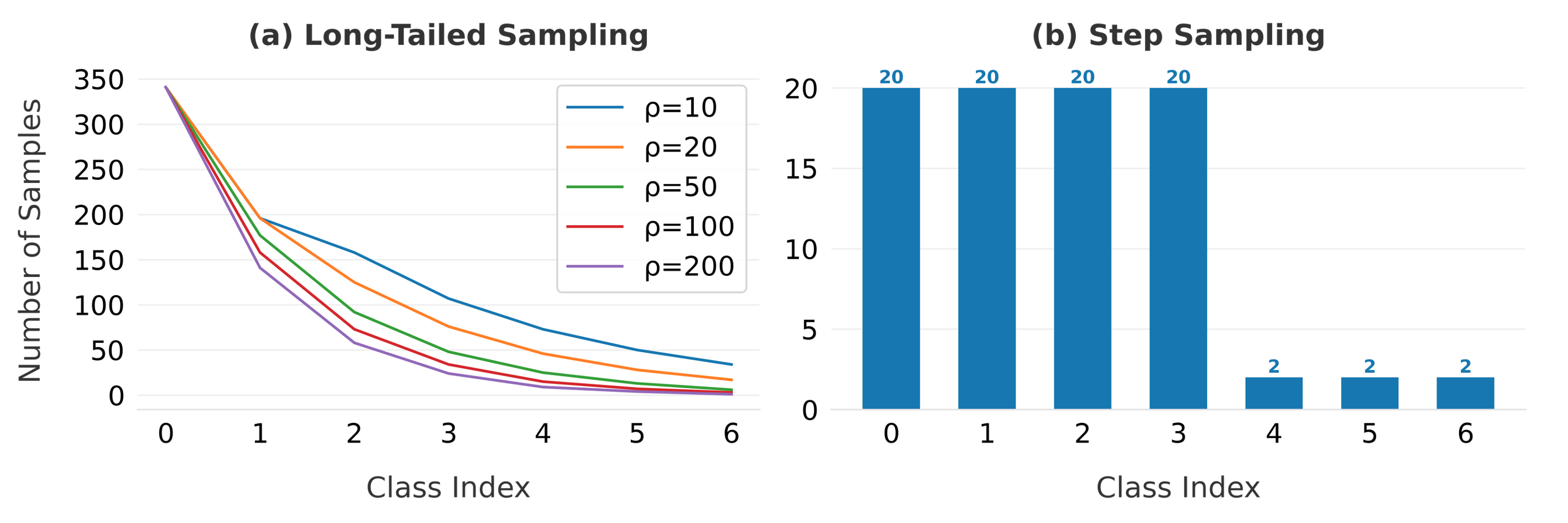}
    \caption{Number of training samples per class in the Cora dataset with (a) long-tailed sampling ($\rho \in {10, 20, 50, 100, 200}$) and (b) step sampling ($\mu=0.5, \rho=10$).}
    \label{fig:sampling}
\end{figure}

To evaluate CILG models under class-imbalanced settings, we may need to create extreme class-imbalanced scenarios not present in real-world datasets. A class-imbalanced train-test split strategy can be used for this purpose: a class-balanced test set is created by randomly selecting the same number of nodes from each class and the remaining nodes assigned to the training set, which will be sampled. There are two class-imbalanced sampling strategies: (1) long-tailed sampling, which corresponds to a supervised node classification scenario, and (2) step sampling, which corresponds to a semi-supervised node classification scenario. Fig. \ref{fig:sampling} compares the number of training samples in each class in the Cora dataset \cite{yang2016revisiting} after long-tailed sampling and step sampling, respectively.

\subsubsection{Long-tailed sampling}
The long-tailed sampling method for node classification is characterized by an imbalance ratio $\rho$. This method was inspired by creating a long-tailed version of the CIFAR dataset in the computer vision domain \cite{cui2019class}. The goal is to maintain the graph connections as much as possible while reducing the number of nodes in each class to follow a long-tailed distribution \cite{park2021graphens}. Typically, the number of training samples in each class decreases exponentially, with a constant factor of $\rho^{1/(K-1)}$, resulting in an exact imbalance ratio of $\rho$.

\subsubsection{Step sampling}
An imbalance ratio $\rho$ and a parameter $\mu$ define the step sampling method, following a well-established semi-supervised node classification setting \cite{yang2016revisiting,kipf2017semisupervised}. A small number $n$ of training samples, e.g., 20, are selected for each class. The method randomly chooses $\mu K$ classes as minority classes and $(1-\mu) K$ classes as majority classes, where $K$ is the number of class labels. From the minority classes, it randomly selects $\rho n$ nodes to form a class-imbalanced training set.

\subsection{Performance Metrics}
We review widely-used performance metrics in CILG research and assess their suitability for evaluating performance under class-imbalanced settings. We consider a binary classification problem, which can be generalized to multiclass classification. A confusion matrix records the correctly and incorrectly predicted instances per class, facilitating the computation of various metrics. True positives (TP) and true negatives (TN) are the numbers of positive and negative samples correctly classified. False positives (FP) and false negatives (FN) represent the misclassified positive and negative samples. Common metrics employed in CILG are outlined below:

\begin{itemize}[leftmargin=*]

    \item \textbf{Acc} (Accuracy) is the ratio of the number of correct predictions to the total number of predictions:
    \begin{equation}
        \text{Acc} = \frac{\text{TP} + \text{TN}}{\text{TP} + \text{TN} + \text{FP} + \text{FN}}.
    \end{equation}
    
    \item \textbf{F$_\beta$} (F-measure) is the weighted average of \textbf{Prec} (precision) and \textbf{Recall}:

    \begin{equation}
        \text{Prec} = \frac{\text{TP}}{\text{TP} + \text{FP}}, \quad
        \text{Rec} = \frac{\text{TP}}{\text{TP} + \text{FN}}, \quad
        \text{F}_\beta = \frac{(1+\beta)^2 \cdot \text{Prec} \cdot \text{Rec}}{\beta^2 \cdot \text{Prec} + \text{Rec}},
    \end{equation}
    where $\beta \in [0, +\infty)$ is a coefficient to adjust the relative importance of precision versus recall.
    
    \item \textbf{bAcc} (Balanced accuracy) is the arithmetic mean of sensitivity and specificity:
    \begin{equation}
        \text{bAcc} 
        = \frac{1}{2} (\text{sensitivity} + \text{specificity})
        = \frac{1}{2} \left(\frac{\text{TP}}{\text{TP} + \text{FN}} + \frac{\text{TN}}{\text{FP} + \text{TN}}\right).
    \end{equation}

    \item \textbf{GM} (Geometric mean) is the geometric mean of sensitivity and specificity:
    \begin{equation}
        \text{GM} 
        = \sqrt{\text{sensitivity} \times \text{specificity}}
        = \sqrt{\frac{\text{TP}}{\text{TP} + \text{FN}} \cdot \frac{\text{TN}}{\text{FP} + \text{TN}}}.        
    \end{equation}
    
    \item \textbf{MCC} (Matthews correlation coefficient) is a correlation coefficient between the actual and predicted binary classifications:
    \begin{equation}
        \text{MCC} = \frac{\text{TP} \times \text{TN} - \text{FP} \times \text{FN}}{\sqrt{(\text{TP} + \text{FP})(\text{TP} + \text{FN})(\text{TN} + \text{FP})(\text{TN} + \text{FN})}}
    \end{equation}
    \item \textbf{AUC} (Area under the curve) calculates the area under the receiver operating characteristic (ROC) curve, relating the true positive and the false positive rate at different classification thresholds.
    
\end{itemize}

\begin{figure}[t]
    \centering
    \includegraphics[width=0.6\textwidth]{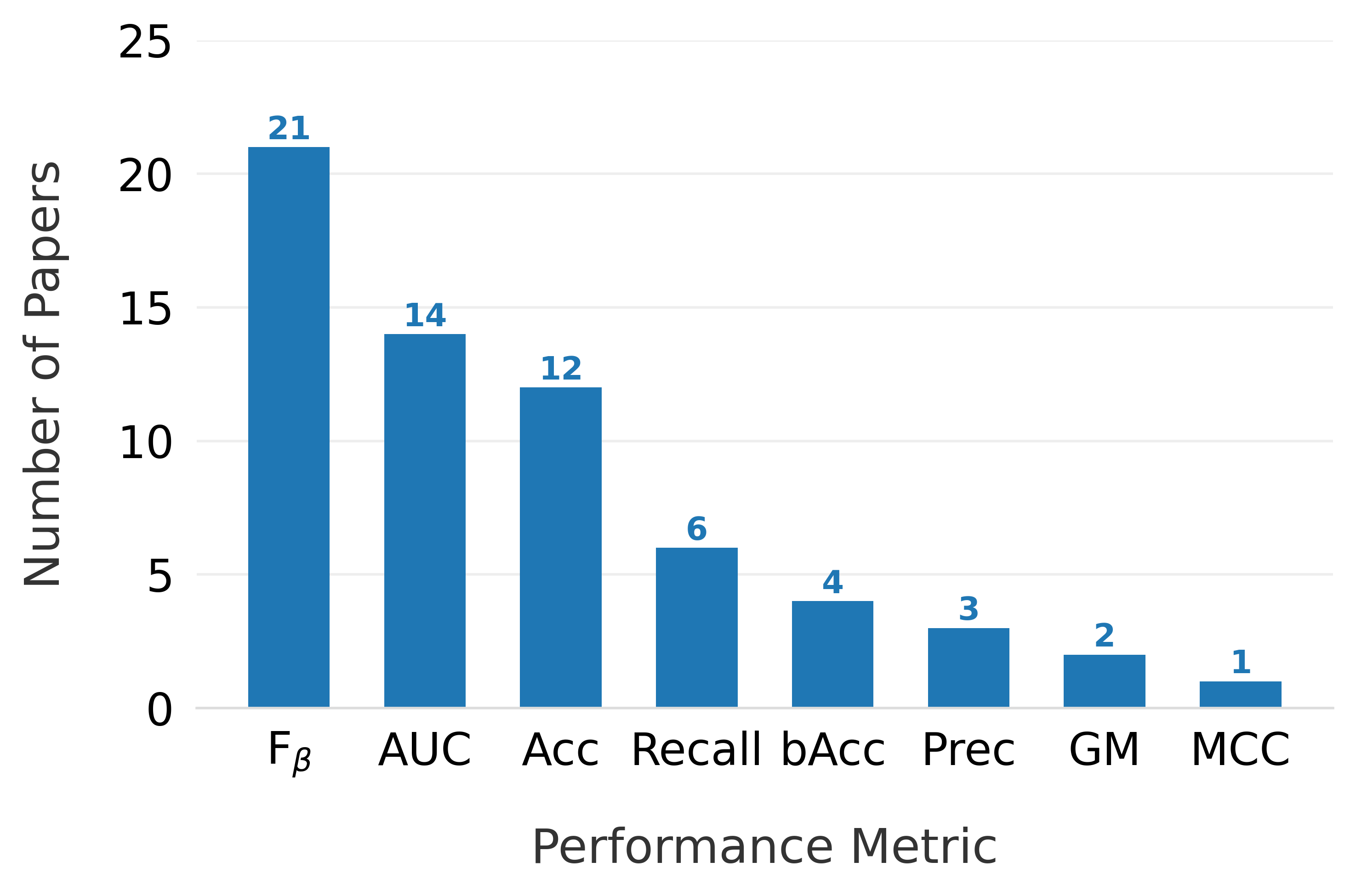}
    \caption{Statistics on the number of CILG papers using each performance metric.}
    \label{fig:metrics}
\end{figure}

Figure \ref{fig:metrics} illustrates the use of performance metrics in CILG papers. Of the 27 existing papers on CILG, nearly half (12) rely on accuracy as the primary metric, with half claiming improved performance based on accuracy results. However, accuracy is known to be biased towards the majority class, failing to distinguish between the number of correctly classified examples from different classes \cite{lopez2013insight}. Furthermore, three out of six papers use precision, and one of three uses recall as a standalone metric, providing a limited perspective on model performance. Consequently, it is crucial for future CILG research to carefully select appropriate evaluation methods that account for potential biases and a more comprehensive understanding of model performance.

\section{Future Directions}
\label{sec:future}
Most existing research on CILG primarily focuses on node classification in homogeneous and homophilous simple graphs. However, it is crucial to investigate and develop effective methods for addressing class imbalance in other graph mining tasks and complex graph types. It is also essential to consider other imbalance scenarios beyond class or quantity imbalances, such as topology imbalances. By broadening the scope of research in these areas, we can enhance the performance of various graph-based applications in real-world situations.

\subsection{Beyond Node Classification}
While most current research on CILG has primarily focused on node classification, exploring and developing effective methods for addressing class imbalance in other graph mining tasks is essential. Examples include edge classification, graph classification, and node regression.

\subsubsection{Edge classification}
Edge classification in class-imbalanced settings involves predicting labels or attributes for edges connecting nodes in a graph while addressing challenges posed by unequal distributions of edge classes. Certain edge classes may be less prevalent in real-world scenarios, making accurate predictions more difficult. 

There are challenges in addressing class-imbalanced edge classification. First, local structures surrounding minority class edges might be less diverse/underrepresented than majority class edges. This can make it difficult for models to capture minority class edges and their distinct characteristics. Second, the interactions between majority and minority class edges may be more complex, with minority edges often forming subgraphs or communities interconnected with majority edges. Capturing and understanding these interactions is crucial for accurate edge classification.

\subsubsection{Graph classification}
Class-imbalanced graph classification aims to predict labels/attributes for entire graphs while dealing with challenges caused by skewed distributions of graph classes.
There are at least two main challenges. First, although it is possible to extract additional supervision for underrepresented nodes from surrounding neighborhoods in node classification, graphs are separate entities. Therefore, aggregating information from other graphs via propagation is not feasible. Second, graph-structured data imbalance can extend beyond the feature/semantic domain, as seen in regular grid or sequence data. In graph data, imbalances may also stem from disparities in a graph topology. Limited training graphs with unrepresentative topologies might poorly define minority classes, making it difficult for models to generalize to those of unseen testing graphs.

\subsubsection{Node regression}
Because research in imbalanced learning has mostly focused on tackling classification problems, addressing these issues in numerical prediction tasks remains limited~\cite{branco2016survey,moniz2017resampling,ribeiro2020imbalanced,oliveira2021biased,silva2022model,yang2021delving}. Regression tasks generally assume equal relevance for all domain values. This assumption is rarely accurate for real-world domains, e.g., finance or meteorology, where the primary objective is often to predict rare or extreme events. Node regression tasks focus on predicting underrepresented numerical target values associated with graph nodes.

A critical challenge in imbalanced node regression is the variability of local graph structure in nodes with extreme target values. These structures might differ significantly from those of nodes with more common target values, making it difficult to understand the factors contributing to extreme target values. Another challenge is ensuring the model's robustness in imbalanced node regression tasks, as it may become overly sensitive to outliers or noise. Techniques to enhance the robustness of the model against such factors are essential for accurate and reliable predictions.

\subsection{Beyond Homogeneous and Homophilous Graphs}
Although a significant portion of the existing research in graph mining has centered on simple graphs that are homogeneous and homophilous, it is crucial to explore the potential for addressing class imbalance in more complex graph types. Heterogeneous, temporal, hypergraphs and heterophilous graphs are four such complex graph types that can also be affected by class imbalance.

\subsubsection{Heterogeneous graphs}
Heterogeneous graphs \cite{wang2022survey} contain multiple types of nodes and edges, reflecting the complex relationships in various real-world scenarios. Addressing class imbalance in heterogeneous graphs presents unique challenges. One such challenge is integrating diverse node and edge types during learning, as the relationships between different entities may vary in strength or relevance. Developing techniques for class-imbalanced heterogeneous graphs could enhance the ability to handle rare user-item interactions and improve the overall recommendation quality in applications such as recommender systems.

\subsubsection{Temporal graphs}
Temporal graphs \cite{holme2012temporal}, also known as dynamic graphs, incorporate the element of time into the graph structure, which can be crucial in understanding how relationships evolve \cite{wang2021modeling,wang2020learning}. Class imbalance in temporal graphs introduces specific challenges, such as accounting for the varying importance of historical information and handling the temporal dependencies between imbalanced classes. Developing methods to address class imbalance in temporal graphs can lead to more accurate predictions and a better understanding of the evolution of relationships over time in domains like social networks, financial networks, or communication networks.

\subsubsection{Hypergraphs}
Hypergraphs \cite{feng2019hypergraph} extend traditional graphs by allowing edges to connect more than two nodes, thus representing higher-order relationships. Class imbalance in hypergraphs presents the challenge of capturing and learning from these higher-order relationships, especially when certain relationships are underrepresented. By addressing class imbalance in hypergraph-based tasks, models can better capture the complex structure of relationships and improve performance in various applications, including bioinformatics and social network analysis.

\subsubsection{Heterophilous graphs}
Heterophilous graphs \cite{zheng2022graph} exhibit variations in the connectivity patterns and attributes of nodes and edges, leading to diverse subpopulations within the graph. In the context of class imbalance, heterophilous graphs present challenges, such as identifying and leveraging the diverse connectivity patterns for learning, as well as addressing the rarity of specific subpopulations or connections. Addressing class imbalance in heterophilous graphs can enhance the accuracy of models for applications such as detecting rare communities in social networks or discovering atypical buyer-seller relationships in e-commerce platforms.

\subsection{Beyond Class and Quantity Imbalance}
In this section, we investigate \emph{topology imbalance} as an extension of the well-established issues of class and quantity imbalance, which refers to the unequal distribution of labeled examples among different classes. Topology imbalance emerges in the context of semi-supervised node classification when the structural importance of labeled nodes in a graph varies, causing some nodes to have a greater impact on the classification of others \cite{chen2021topology}.
This imbalance can also appear when certain topological motifs (\emph{i.e.,} patterns of connections between nodes) in a graph are unevenly represented in the training data \cite{zhao2022topoimb}. Consequently, Graph Neural Networks (GNNs) might encounter difficulties learning a generalizable classification boundary when specific topological patterns are under-represented, allowing certain patterns to dominate the learning process.

Another perspective on topology imbalance involves examining the global distribution of supervision information, considering factors such as under-reaching and over-squashing \cite{sun2022position}. Under-reaching happens when the influence of labeled nodes diminishes with increasing topology distance, causing nodes far from labeled ones to receive insufficient supervision information. Conversely, over-squashing occurs when valuable supervision information from labeled nodes gets compressed as it passes through narrow paths, mixed with other less relevant information, ultimately degrading the supervision information quality received by specific nodes.

\section{Conclusion}
In conclusion, this paper presents the first comprehensive survey on class-imbalanced learning on graphs (CILG), addressing the significant challenges of class imbalance in graph data. By systematically categorizing existing CILG methods, discussing pressing research questions, and offering a roadmap for future investigations, the paper aims to raise awareness within the graph machine learning community and encourage further research in this emerging area. As the applicability of class-imbalanced learning techniques expands across various graph-based domains, researchers and practitioners will be better equipped to develop effective and efficient solutions for handling class imbalance problems, benefiting a wide range of applications and industries.


\bibliographystyle{ACM-Reference-Format}
\bibliography{references}

\appendix

\end{document}